# A Review on Digital Pixel Sensors

Md Rahatul Islam Udoy[1], Graduate Student Member, IEEE, Shamiul Alam[1], Graduate Student Member, IEEE, Md Mazharul Islam[1], Graduate Student Member, IEEE, Akhilesh jaiswal[2], Member, IEEE and Ahmedullah Aziz[1], Senior Member, IEEE
[1]Department of Electrical Engineering and Computer Science, University of Tennessee, Knoxville, TN 37996 USA.
[2]Department of Electrical and Computer Engineering, University of Wisconsin-Madison, WI 53706 USA

Corresponding author: Ahmedullah Aziz (e-mail: aziz@utk.edu).

**ABSTRACT** Digital pixel sensor (DPS) has evolved as a pivotal component in modern imaging systems and has the potential to revolutionize various fields such as medical imaging, astronomy, surveillance, IoT devices, etc. Compared to analog pixel sensors, the DPS offers high speed and good image quality. However, the introduced intrinsic complexity within each pixel, primarily attributed to the accommodation of the ADC circuit, engenders a substantial increase in the pixel pitch. Unfortunately, such a pronounced escalation in pixel pitch drastically undermines the feasibility of achieving high-density integration, which is an obstacle that significantly narrows down the field of potential applications. Nonetheless, designing compact conversion circuits along with strategic integration of 3D architectural paradigms can be a potential remedy to the prevailing situation. This review article presents a comprehensive overview of the vast area of DPS technology. The operating principles, advantages, and challenges of different types of DPS circuits have been analyzed. We categorize the schemes into several categories based on ADC operation. A comparative study based on different performance metrics has also been showcased for a well-rounded understanding.

## I. INTRODUCTION

Vision sensors have emerged as fundamental components that create a bridge between the digital and physical worlds in the rapidly evolving landscape of including but not limited to artificial intelligence (AI), robotics, space observation, and computer vision [1]–[4]. With the crucial input from vision sensors, autonomous vehicles achieve the essential capability to perceive and interpret their surroundings and thus act as the 'eye' of the vehicle, providing the vital data that fuels the decision-making process [5], [6]. Using vision sensors, robots can capture visual data and employ advanced algorithms, which enable them to do complex tasks and enhance safety[7]. This fosters seamless integration of robots in various industries [8]. Precise vision sensors play an important role in the medical field[9], [10]. They are often integrated into complex machines such as surgical robots, enabling surgeons to navigate delicate anatomy and perform intricate tasks with minimal invasion [11]–[15]. Targeted drug delivery for precision medicine is also benefited from the development of compact image sensors [16]. ML algorithm driven medical imaging holds the potential to redefine the landscape of medical diagnostics with a level of precision and reliability previously unachievable [17]. ML based image sensors bring a paradigm shift in the field of surveillance [18]. Improvement in image sensors can elevate the performance of various types of detection tasks such as face recognition, object detection, digit, and text recognition, etc. [19]–[23].

A camera can be analogized to the biological eye [24], [25] because of the resemblances present among their respective components (shown in Fig. 1). The main component of the camera is the image sensor that converts light to digital signals. The researchers have tried numerous ways to design image sensors. Charge coupled devices (CCD) are one of the early-stage sensors [26], [27], invented back in 1970. These sensors did not continue because they were very slow, required several clock-cycles for completing a readout, and the situation deteriorates with the increasing number of pixels [28]. CMOS image sensors took the place of CCDs by overcoming the mentioned issues. However, new issues arose for CMOS image sensors such, as low fill factor, low sensitivity, noise, low dynamic range (DR), and lower image quality than CCD [29]. The research is still ongoing to overcome those issues. The first CMOS image sensor was the passive pixel sensor (PPS), which consists of only a photodetector and a transistor [30]. PPSs could not take off massively because they suffered from high noise levels due to the mismatch between the large bus capacitance and low pixel capacitance [31]. To mitigate this issue, active pixel sensors were introduced (Fig. 2). The first-generation active pixel sensors produce analog voltage at the pixel output [32] and can be termed as analog pixel sensors. The ADC circuitry lies at the end of each column [33]. The analog voltage generated from the pixel travels at the end of the column and is converted to digital signal [34]. This kind of design exhibits some issues such as low speed, high noise, etc. and a detailed discussion about these issues is presented in section III [35], [36]. As a remedy, a new kind of image sensor was suggested, where an ADC circuit resides inside each pixel [37], [38]. These are called digital pixel sensors (DPS) and the ADCs are known as pixel-parallel ADC or in-pixel ADC. However, in the case of DPS, the new issue is the bulky pixel circuit, which hinders the way of making high-density pixel arrays, therefore making compact cameras with DPS becomes impractical. In order to address this problem, relentless effort has been invested and continues to persist currently. Researchers are trying several types of circuits to design DPSs with desired characteristics and this review article delivers a comprehensive idea of those approaches. The analysis shown throughout this article, along with a comparison among the approaches (presented in section VII), will aid in forming an understanding of the trajectory that this research trend is taking.

In this review article, we articulate the discussion in the following way- Section II will build knowledge on the basic pixel circuit working principle. In section III, we discuss the incentives behind opting for digital pixel sensors. Section IV starts with categorizing different circuit design approaches and then presents a detailed study of different DPSs. In section V, the 3D integration technique and an application of this architecture are presented. In section VI, a noise analysis of DPS is presented. Section VII demonstrates the juxtaposition of different DPSs provided through a comparative analysis. Section VIII explores applications of DPS. Section IX offers insights into current trends and outlines future directions in DPS technology, providing a comprehensive understanding of its evolving landscape.

## II. BASIC PIXEL CIRCUIT MECHANISM

The basic 3-transistor (3-T) pixel circuit diagram is depicted in Fig. 2(a). It consists of 1 reverse-biased photodetector and 3 transistors. The three transistors $M_1$, $M_2$, and $M_3$ are called reset, source follower and row selector, respectively. Fig. 2(b) shows the timing diagram of a basic pixel circuit. The circuit operation begins with turning on the reset transistor, which resets the photodetector (PD) node voltage ($V_{PD}$) to $V_{RESET} - V_{TH}$ (when a soft reset is used) or to $V_{RESET}$ (when a hard reset is used or a PMOS is used as reset transistor [39]. However, PMOS transistors occupy a larger area in the layout). Here, $V_{TH}$ is the threshold voltage of the reset transistor. After resetting, the integration period begins with turning off the reset transistor. Photocurrent due to photoelectrons generated in the photodetector will cause a voltage drop at the node PD according to the following equation-

$$\frac{dv}{dt} = \frac{I_{PD}}{C_{PD}} \qquad (1)$$

Here, $I_{PD}$ is the photocurrent and $C_{PD}$ is the internal capacitance of the photodetector. Higher illumination causes higher $I_{PD}$ which results in faster drop of $V_{PD}$, which is shown in the second cycle of Fig. 2(b). The source follower transistor $M_2$ is used to transfer the signal of the PD node without disrupting the accumulated charge. The last transistor $M_3$ of the basic pixel circuit, is used for selecting the pixel to read out from the array. The output signal (the amount of $V_{PD}$ drop) of this pixel is analog.

In a digital pixel circuit, digitization starts from the PD node and ADC is implemented inside the pixel circuit. The basic concept of a digital pixel is shown in Fig. 2(c). There are numerous types of digitization techniques in a pixel circuit, which will be discussed in the subsequent sections.

There are some commonly used terminologies in the field of digital image sensors, such as-

Dynamic Range (DR): It measures the ability to detect the light of the lowest and the highest possible luminance [40]. It can be described as the maximum possible swing of photo-detection of the image sensor [41]. When this swing is calculated among the pixels of the whole array for a single frame, it is called the intra-scene DR [42]. The DR can be defined as the following equation-

$$DR = 20 \log_{10} \frac{I_{max}}{I_{min}} \qquad (2)$$

Here, $I_{max}$ ($I_{min}$) is the highest (lowest) possible non-saturated photocurrent for the brightest (dimmest) light. The unit of dynamic range is dB. Although there are subtle differences among luminance, light intensity, illumination level and brightness, we will generally use them in this review to indicate the amount of light falling on the sensor.

DR of natural scenes can be measured using the following formula:

$$DR = 20 \log_{10} \frac{L_{max}}{L_{min}} \qquad (3)$$

Here, $L_{max}$ and $L_{min}$ are the luminance (measured in lux) of the brightest light and the dimmest light, respectively. Usually, the brightest and the dimmest light of a natural scene are $10^{-3}$ lux and $10^5$ lux, respectively [43], which results in a dynamic range of 160 dB.

Full-well capacity: It refers to the maximum charge that can be accumulated in the photodiode node [44]. This charge ($Q_{max}$) can be defined as the following equation-

$$Q_{max} = V_{PD} \times C_{PD} \qquad (4)$$

Large well capacity is favorable for high DR [45].

Fill Factor (FF): It can be expressed as the ratio of the photo-sensitive area ($A_{Ph}$) to the total pixel area ($A_{pix}$) and presented in a percentage form.

$$FF = \frac{A_{Ph}}{A_{pix}} \times 100\% \quad (5)$$

For making an image sensor better, the FF should be higher [46]. The total pixel area is dependent on the complexity of the readout and ADC circuitry. A simple circuit with a lower number of transistors can reduce $A_{pix}$, which improves FF. 3D integration can also improve the FF rapidly, which will be explained in section V.

Dark Current: When the pixel is in full darkness, there should be no photocurrent, but due to room temperature, a certain amount of electron-hole pair generation continues and that generates a specific level of photo-current, which is called the dark current [47]. This current determines the $I_{min}$ of the DR (i.e., the floor of the swing). It is clear from equation (2) that the lower the dark current, the better the DR.

The dark current is not uniform for every pixel throughout the array due to inconsistent thermal noise levels. This non-uniformity noise is called the dark signal non-uniformity (DSNU) [48].

Sensitivity and Responsivity: The detector sensitivity of a photodiode refers to its capability to convert incoming light or photons into an electrical current or voltage. This sensitivity is usually expressed in terms of the photodiode's responsivity (R), which measures the photocurrent or voltage produced per unit of incident optical power ($P_{in}$) [49].

Signal-to-Noise Ratio (SNR): This is defined as the ratio of the power due to the photodetector current ($I_{PD}$) to the input referred noise power [50], [51].

$$SNR = \frac{I_{PD}^2}{\Sigma(I_{noise})^2} \quad (6)$$

Here, $I_{noise}$ currents are the equivalent currents due to different noises. SNR is usually measured in dB.

III. WHY DIGITAL PIXEL SENSOR?

Digital pixel sensors offer several advantages over its counterpart, analog pixel sensor. In the latter one, the analog output from the pixel is converted to digital at the end of the column. The voltage swing decreases due to the threshold voltages of the transistors in Fig. 2(a)) [28], [29], [52]. As a consequence, with the technology scaling, the output voltage swing of the APS is drastically constrained because the reduction in supply voltage outpaces that of the threshold voltage [53]. Dynamic range is directly tied to the voltage swing. If the voltage swing decreases, the difference between the brightest and darkest signals the sensor can accurately capture diminishes. Furthermore, the signals traversing the pixel array are analog voltages (Fig. 3(a)), making them susceptible to noise coupling through either the power lines or the substrate [54]. Whereas, in the case of DPS, the signals run through the long pixel array are digital, hence immune to noises.

Digital pixel sensors offer better performance in the case of high-speed imaging, compared to APS. For the latter one, the single digitization module at the end of the column has to handle the analog to digital conversion of all the pixels ($m$ number of pixels in a $m \times n$ pixel array) in the column (Fig. 3(a)). The ADC has to convert the pixel voltages one by one, thus imposing a speed constraint. It is obvious that, with increasing the size of the array, this issue escalates. On the other hand, DPS architecture mitigates this issue by integrating an ADC within each individual pixel (Fig. 3(b)). This design allows each pixel to perform its own analog-to-digital conversion independently. As a result, the conversion process is highly parallelized, with each pixel simultaneously converting its analog voltage to a digital signal. This parallelism significantly enhances the speed of the imaging process, as there is no longer a single ADC module acting as a bottleneck.

The relative motion between the camera and the objects of a scene causes visual artifacts like skewed or wobbly objects in analog pixel sensors, which is called the rolling shutter effect. This is because all pixels in a column share an ADC and thus cannot be read out simultaneously. So, a row-by-row readout operation takes place and there are readout time differences between two rows in a single frame (Fig. 4(a)). This causes distortion in capturing moving objects as depicted in Fig. 4(b), where the shape of the blades of a helicopter rotor appears skewed in the image, even though in real life, they are straight in shape. In the case of digital pixel sensors, as the pixels do not share ADC, they can be read out simultaneously. When the entire pixel array is read out in a single instant, it is called global shutter.

Moreover, unlike the APS, random noise in a DPS does not comprise readout noise, column amplifier noise, source follower noise, etc., because of pixel-level conversion and discarding several noisy transistors such as source follower, column amplifier transistor, etc. Although reset noise and shot noise is still prevalent in the DPSs [55].

Fixed pattern noise (FPN) is quantified without any incident light on the sensor. This noise is not uniform throughout the array (i.e., it varies from pixel to pixel). Pixel and column FPN come from the source follower and the column amplifier, respectively. DPS eradicates these noises from their origins by simply not having the origins [55].

The above-mentioned advantages of DPSs come with costs. Digital pixel sensors (DPS) exhibit disadvantages primarily in terms of increased pixel size, complexity, manufacturing costs, and higher power consumption compared to analog pixel sensors (APS). Each pixel in a DPS requires its own ADC, leading to larger pixel sizes to accommodate this additional circuitry. This larger size can limit the overall spatial resolution of the sensor array, affecting applications that demand high pixel density. The integration of ADCs at the pixel level also results in reducing yield rates. This occurs because the complexity increases the chances of defects or errors in the manufacturing process, thereby reducing the overall yield of functional sensors. Furthermore, the operation of multiple ADCs simultaneously consumes more power compared to APS. These factors make DPS less efficient in terms of size, cost, and power consumption, particularly in applications where compact size, low cost, and energy efficiency are critical considerations.

However, Designing compact ADC circuits while maintaining high ADC resolution and implementing 3D integration techniques represent promising solutions to mitigate the challenges posed by integrating ADCs at the pixel level in digital pixel sensors (DPS). Compact ADC designs focus on minimizing the physical footprint of the ADC circuitry within each pixel, thereby reducing the impact on pixel size and overall sensor resolution. High ADC resolution ensures that each pixel can accurately convert analog signals to digital data, crucial for maintaining image quality and sensor performance. Additionally, 3D integration techniques involve stacking multiple layers of electronic components vertically, enabling more efficient use of space and potentially reducing manufacturing complexities (We discuss 3D integration in section V). These advancements not only help address size and yield rate concerns [56], but also contribute to enhancing the overall efficiency.

## IV. CLASSIFICATION

A classification of CMOS image sensor is illustrated in Fig. 5. When a pixel circuit contains at least an active circuitry at the pixel level, it is called an active pixel sensor. On the contrary, a passive pixel circuit contains only a photodetector and a transistor switch [30]; This kind of pixel sensor is obsolete now because of suffering from slow readout and high noise. Active pixel sensors can be classified based on where the ADC is taking place and which domain is being used for the ADC conversion. When the output from the individual pixel is in analog form, then it can be termed as analog pixel sensor. In some earlier versions of analog pixel sensor-based cameras, the ADC operation takes place off-chip, which is outdated now. Most modern cameras have converters inside the chip. In some cases, there is only one ADC circuitry assigned for the whole pixel array, this is named as single ADC in the classification. In the column parallel ADC, at the end of each column of the pixel array, there is an ADC [33].

The focus of this review is on the digital pixel sensors (DPS) where the whole or maximum portion of ADC circuitry must reside inside the pixel circuit and hence, they are also called pixel-level ADC [37], [38]. DPSs can be categorized based on which domain is being used for conversion. In voltage domain conversions, the pixel voltage is directly quantized. Several literature used voltage domain conversion [57], [58], out of which, two DPSs will be discussed in sections IV(A) and IV(B). They are- the multichannel bit serial (MCBS) and the successive approximation register (SAR) ADC. On the other hand, in the case of time domain conversion, the pixel voltage level is converted to pulses or spikes. Delta-sigma converters [59] and pulse frequency modulation (PFM) [60] based sensors count the number of pulses in a certain amount of time. The decimator, an important and area-hungry part of the delta-sigma converter, can reside inside the pixel (in-pixel decimation) [59] or can be shared among all the pixels in a column or a pixel block (off-pixel decimation) [61]. The pulse width modulation (PWM) technique counts the time to get the first spike[62].

In some literature, the voltage domain and the domain conversions are merged to achieve the advantages of both. Triple quantization is one of the best example of this, in which three modes of quantization are applied for a single exposure [63] (details in section IV(F)).

### A. MULTI-CHANNEL BIT SERIAL (MCBS)

In analog to digital conversion, usually, an analog voltage is converted to a binary code of a certain number ($m$) of bits, which requires $m$ bit memory inside each pixel. This $m$ bit memory takes a huge chunk of space of a pixel. Yang et al. showed a mechanism to generate each bit independently and combine $m$-bits sequentially; this is called bit-serial operation [58]. The key advantage of bit-serial operation lies in its requirement of only a 1-bit memory per pixel, in contrast to the m-bits required in traditional ADC architectures. This substantial reduction in memory requirements translates directly into significant area savings within each pixel. By minimizing the amount of memory needed to store digital values, bit-serial ADCs optimize the physical layout of pixels on a sensor, allowing for higher pixel density. Table I shows an example of 3-bit quantization, where it is assumed that the input voltage range is between 0-1 V. For a certain input voltage, whether the LSB will be encoded with 1 or 0 is decided by asking the question, does the input voltage lie between 0.125-0.375 V or 0.625-0.875 V? If yes, then the LSB is 1, otherwise 0. In a similar fashion, other bit positions are also determined.

Fig. 6(a) shows the block diagram of multichannel bit serial operation. Each channel has a comparator and latch pair. The comparator takes analog pixel voltage in its positive input terminal. The controller block and the digital to analog converter (DAC) block are shared among the channels. The controller generates gray code word and the single bit (BITX) that has to be latched. The code word is converted to a voltage level in the DAC block according to table I and this voltage is sent to the negative terminal of the comparator. In this way, DAC generates the boundary points of different voltage ranges shown in the 1st column of Table I, and the comparator performs multiple comparisons to determine the voltage range in which the input pixel voltage lies. For example, to decide whether the LSB is 1 or not, the comparator has to compare the analog pixel voltage four times with four boundary points from Table I (0.125, 0.375, 0.625 and 0.875). In a similar way, for making decision on the middle bit, two comparisons have to be made for two boundary points (0.25 and 0.75). And for the MSB, only one comparison is required (greater than 0.5 or not). Thanks to the gray coding, that helps to keep the number of comparisons lower.

After determining a bit of a certain position and performing the latching operation of that bit, a read-out operation takes place to extract the bit from each channel. This generates a bit-plane for the whole pixel array. And sequentially more bit planes are formed for other bit positions shown in Fig. 6(b). For digitization with *m*-bit depth, *m* number of bit planes must be generated which form one frame. At a particular coordinate of the pixel array, the bits generated along the *m* number of bit planes represent the digitized value of that coordinate. Frame generation with this kind of approach opens several opportunities such as applying in-sensor ML algorithm, region of interest (ROI) windowing, etc.

Traditional ADCs often face challenges related to area constraints, power consumption, and speed limitations. However, the innovative approach of generating each bit independently and sequentially combining them addresses these challenges. The utilization of comparators and latches, along with a centralized controller and digital-to-analog converter (DAC) block, ensures an accurate determination of voltage ranges for each input signal, while the integration of gray coding minimizes computational complexity.

The idea of MCBS ADC is later extended to enhance the dynamic range by leveraging multiple sampling with exponentially increasing exposure times *T, 2T, 4T*, and so forth, up to $2^qT$ [64]. At each of the sampling times, the pixel output is digitized to *p* bits. At the end of time $2^qT$, each pixel value is digitized to *p+q* bit binary number, offering floating-point resolution. The binary number is then converted to a floating-point number with a *p* bit mantissa and an exponent spanning from 0 to *q*, which increases the DR by a factor of $2^q$. To better understand the idea, the following example can be observed. Suppose a scene is sampled three times (*T, 2T, and 4T*), and at each time, the pixel value is digitized with a resolution of 2 bits. So, in this case, *p*=2 and *q*=2. Let's explore the scenario where three unique levels of illumination ($L_1$, $L_2$, and $L_3$) are cast upon three individual pixels (shown in Fig. 6(c)). At sampling time *T*, the pixel output is digitized to $b_1b_2$ (00 for $L_1$, 01 for $L_2$, and 11 for $L_3$). Also, at sampling time *2T,* the pixel value is digitized to 2 bits. As the second sampling time is twice the first, the two bits in this case are $b_2b_3$ (if the pixel output is not saturated to the saturation voltage, $V_{sat}$) or 11 (if the pixel output is saturated). In both scenarios, the ADC is only required to generate the LSB of $b_2b_3,$ which is $b_3$ (0 for $L_1$, 0 for $L_2$, and 1 for $L_3$). In a similar fashion, at sampling time *4T,* only the bit $b_4$ has to be generated. Table II shows how these 4-bit numbers can be converted to a floating-point number representation with a mantissa and an exponent.

B. SUCCESSIVE APPROXIMATION REGISTER (SAR)

A successive approximation register (SAR) type analog-to-digital converter (ADC) is used to convert analog pixel signals into digital data by operating through a process of iterative approximation. The SAR ADC compares the input voltage against a reference voltage and progressively refines its digital output code until it accurately represents the analog input. This refinement occurs within a register, where the ADC's digital representation of the analog signal is manipulated. Utilizing a comparator network, the ADC compares the input voltage with the output of a digital-to-analog converter (DAC), adjusting the digital output code bit by bit until convergence is achieved. Fig. 6(d) illustrates the basic block diagram of a SAR ADC[57]. It comprises five essential blocks, they are sample and hold (S/H), comparator, control logic, registers (SAR), and digital to analog converters. Analog pixel voltage is inserted into the positive terminal of the comparator. Control logic changes the bits in registers according to the sequence of operation. DAC generates proper reference voltage for the comparator using the information of register output states. To understand the basic mechanism of a SAR type ADC, we can take an example of 3-bit resolution (8 quantization levels). Suppose the analog pixel voltage range is 0-1 V and the reference voltage ($V_{REF}$) is 1 V. The operation starts with setting the register output to 100 (i.e., MSB=1) and DAC generates $V_{REF}/2$ (As 100 is in the midway of the 8 quantization levels), which is 0.5 V in our example. DAC compares the analog input from the pixel with this 0.5 V, If the input is greater than 0.5 V, the control logic block keeps the MSB of the register output to 1, otherwise makes it 0. Now the ADC moves to the next most significant bit (NMSB) and sets it to 1, which means the new register output is 0**1**0 or 1**1**0 (shown in Fig. 6(e)). Now DAC generates the corresponding voltage ($V_{REF}/4$ if the register output is 010 or $3V_{REF}/4$ if the register output is 110) which will be compared with the input again. In this way, the approach continues all the way to the LSB, and the final register output is the approximate digital representation of the analog input. This entire operation can be represented as a decision tree shown in Fig. 6(e). The resolution limit of the SAR ADC is mainly set by the DAC thus it cannot achieve very high resolution.

C. DELTA-SIGMA ADC

Delta-Sigma ADC utilizes oversampling and noise shaping to achieve high-resolution and high SNR digitization. This type of ADC has the following functional blocks- a difference amplifier, an integrator, 1-bit ADC (which is basically a comparator), 1-bit DAC, a digital filter, and a decimator. The block diagram is illustrated in Fig. 6(f)[65]. The analog pixel voltage is sampled at a very high frequency (usually multiples of Nyquist rate), which is known as oversampling and the ratio of the oversampling frequency to the Nyquist frequency is called the oversampling ratio (OSR). The difference amplifier generates the difference between the oversampled analog input signal and the analog version of the quantized output signal generated by the DAC. This difference represents the quantization error, which is then sent to the integrator. After integration, the signal is sent to the comparator, which generates 1 or 0 by comparing with a reference voltage. This process is iterated several times to generate a bit-stream at the modulator output. In the bit-stream, the density of ones in a certain amount of time is proportional to the level of analog pixel voltage.

Oversampling spreads the quantization noise over a larger frequency range (Fig. 6(g)). Noise shaping shifts the noise to high frequency. This improves the SNR significantly. The modulator output goes through a digital filter that discards the quantization noise of high frequency region. Then the filtered signal is decimated or slowed down to the Nyquist rate. Decimator reduces data rate by averaging a certain of bits. The slowed down data is the final representation of the analog pixel voltage. Unlike the SAR type, the resolution of delta-sigma ADC does not depend on any analog components such as DAC or comparator. The resolution can be adjusted by tuning the over-sampling ratio. A digital-to-analog converter (DAC) is a crucial component in the delta-sigma ADC. Operating on the principle of binary-weighted resistors or pulse-width modulation, DACs take digital input in the form of binary code and produce continuous analog output proportional to the digital input. In binary-weighted DACs, each bit of the digital input corresponds to a specific resistor, and the voltage output is determined by the weighted sum of these resistors. In pulse-width modulation DACs, the digital input controls the width of pulses in a continuous stream, where the average value of these pulses represents the analog output.

Usually the decimator is area hungry, hence sometimes the decimator is kept outside the pixel and shared by multiple pixels or all the pixels of a column [66] or at the chip-level [61], [65]. This type of arrangement is called off-pixel decimation. But the data rate before the decimation is high and the pixel array has to accommodate that high data rate which limits the frame rate, resolution, etc. On the contrary, when each pixel has its own decimator inside, it is called in-pixel decimation. The decimators are kept inside the pixel to get the advantage of a true digital pixel such as reduced output data rate from the pixels of an array. But to accommodate the decimator inside the pixel, it has to be very compact. Mahmoodi et al. designed the decimator with a minimized area and fitted reasonably inside the pixel [59]. According to the layout of their chip, the decimation filter occupied almost half of the whole ADC area.

D. PULSE FREQUENCY MODULATION (PFM)
As the name suggests, when the frequency of the generated pulses is modulated by the amplitude of the input analog pixel voltage, then it can be categorized as the pulse frequency modulation (PFM) or photocurrent to frequency converter (PFC). A representative figure of this type of converter and working principle is shown in Fig. 7. PFM based pixel circuit consists of a reset transistor, a photodiode, a comparator, a counter, and a feedback block (Fig. 7(a))[60]. The operation starts with enabling the reset transistor to insert voltage $V_{RESET}$ to the $PD$ node which is connected to the positive input terminal of the comparator. The other input terminal of the comparator is connected to a predefined reference voltage $V_{REF}$. At the very beginning after resetting, as $V_{PD} (= V_{RESET}) > V_{REF}$, the comparator generates a positive output which will disable the PMOS reset transistor via feedback block. Now, the integration period starts and when the light falls on the photodiode, photocurrent is generated, $V_{PD}$ starts to drop (Fig. 7 (b), (e)).

It can go down to $V_{REF}$ and the rate of dropping can be expressed using the equation (7) [67].

$$\frac{V_{RESET} - V_{REF}}{T_{PD}} = \frac{I_{PD} + I_D}{C_{PD}} \quad (7)$$

Here $T_{PD}$ is the time to fall $V_{PD}$ from $V_{RESET}$ to $V_{REF}$. $I_{PD}$ and $I_D$ are the photogenerated current and dark current respectively. This equation simply emerges from the equation of voltage and current of the $PD$ node capacitance. In case of high illumination, $I_{PD}$ will be high, which implies it will take less time to drop $V_{PD}$ (Fig. 7 (b)). And in case of lower illumination, it will take more time (Fig. 7(e)). At the moment $V_{PD}$ crosses $V_{REF}$, the comparator flips and turns on the PMOS reset transistor; this is called self-resetting. This will again set $V_{PD}$ to $V_{RESET}$ and the same cycle goes on. This process continues, the comparator flips whenever the $V_{PD}$ crosses $V_{REF}$ and generates pulses or spikes that many times. This spike generation can be compared with the biological vision in which generated spikes are used to realize the picture. The number of pulses ($N_{PULSE}$) generated in a time $T_{INT}$, will depend on the time $T_{PD}$ (i.e., the illumination level), which can be shown in equation (8).

$$N_{PULSE} = \frac{T_{INT}}{T_{PD}} = \frac{(I_{PD} + I_D)}{(V_{RESET} - V_{REF}) \times C_{PD}} \times T_{INT} \quad (8)$$

Fig. 7(c) and (f) show the output pulses in the same amount of time ($T_{INT}$) for high and low illuminations respectively. This number of pulses is counted in a counter and the count value is latched in a memory, then the counter is reset for the next operation. This count value represents the digital value of the analog pixel voltage. Fig. 7(d) and (g) show the shape of the output pulses for a slightly different circuit topology where the $V_{PD}$ and $V_{REF}$ are connected to the negative and positive terminals of the comparator respectively. For a certain level of illumination and integration time, both topologies generate the same number of pulses.

Chen et al. designed a PFM with in-pixel variable reference [68]. The circuit diagram is shown in Fig. 8 (a). In addition to the common PFM circuit, they added two inverters and a reference capacitor to generate the variable reference. When the comparator output is high, $X_P$ turns on and charges up the capacitor to increase $V_{REF}$ gradually. On the other hand, when the comparator output is low, it turns on the $X_N$ and discharges the capacitor. The timing diagram in Fig. 8(b) and (d) show the variable $V_{REF}$ and the $V_{PD}$ for high and low illuminations respectively. The pulse generation mechanism is the same as the normal fixed reference voltage PFM. Fig. 8(c) and (e) show the generated pulse at the output of the comparator. The advantage of gradually increasing reference voltage (to near the $V_{REF}$) becomes prominent in the low light condition. To better understand this advantage, we can observe Fig. 8(d)-(g). For a light of certain low luminance, the increasing $V_{REF}$ can generate an output pulse (Fig. 8(d) and (e)) whereas the PFM with a fixed reference cannot generate any output. So, the increasing reference voltage can extend the range of minimum detectable light, which increases the dynamic range.

The best advantage of PFM based pixel circuit is that the ADC resolution does not depend on the full well capacity of the photodetector. Rather, it uses the full well capacity multiple times in a time range ($T_{INT}$). The full well capacity depends on the highest voltage level that can be used as $V_{RESET}$. And this highest possible voltage level in a chip is decreasing dramatically with scaling. So, keeping this constraint in mind, the PFM could be a good choice over voltage-domain based DPSs.

### E. PULSE WIDTH MODULATION (PWM)

PWM based technique uses the simpler part of the theory of PFM. It generally converts the time to first spike (the time needed for $V_{PD}$ to reach $V_{REF}$ from $V_{RESET}$) to a digital value. Fig. 9(a) shows the circuit diagram of a PWM based pixel circuit [69]. This is similar to the PFM circuit. The difference is that the output of the comparator is connected to a memory, which is used to store the count value of the down-counter (up-counter can also be used). The operation of the circuit starts with enabling the reset transistor, which sets $V_{PD} = V_{RESET}$. Then the reset transistor is turned off, the counter starts counting, the $V_{PD}$ starts to fall as the photodiode is illuminated (Fig. 9(b)). When the $V_{PD}$ crosses $V_{REF}$, the comparator flips, generates a pulse (Fig. 9(c)), enables the reset transistor (self-resetting) and enables the memory to latch the count value of the counter which represents the time of dropping $V_{PD}$ from $V_{RESET}$ to $V_{REF}$ (Fig. 9(d)). Higher count value indicates higher illumination (lower illumination) in the case of a down counter (up counter). Counters play a vital role in PWM-based pixel sensors. Typically constructed using flip-flops or other sequential logic elements, a counter circuit increments or decrements its count in response to clock pulses. Each count represents a unique binary value, allowing the circuit to progress through a predetermined sequence. The output of the counter circuit is often represented in binary form, with each flip-flop representing a bit of the count.

Unlike the PFM, in the case of PWM, a pixel sets its own integration time which results in a wider dynamic range. As PWM requires a single output pulse for a conversion, it reduces the dynamic power consumption and switching noise [70]. But unfortunately, according to equation (1), the pulse width or time is inversely related to the photocurrent ($I_{PD} \alpha \ 1/T$), which makes a non-linear relationship. Two quantization methodologies are employed for sampling photocurrent using counter circuits. The first method, known as uniform quantization (UQ), involves uniformly sampling the time domain. In contrast, the second method selects quantization times to ensure uniform sampling of the resolved photocurrent, resulting in non-uniform time domain quantization (NUQ) but uniform steps in photocurrent [70].

### F. TRIPLE QUANTIZATION (3Q)

Liu et al. introduced a sequential triple quantization (3Q) mechanism in a single exposure that allows an ultra-wide dynamic range[63]. The three quantization modes are- high gain mode for low illumination, low gain mode for mid-range illumination and the time-to-saturation (TTS) mode for bright lights. They have also used 3D integration where the whole circuit is distributed in different layers to allow a high-density pixel array. The first layer consists of six transistors, one photodiode and a capacitor. The six transistors are- reset (RST), dual conversion gain (DCG), antiblooming (AB), transfer (TG), source follower (SF) and pixel bias (PB) transistor (Fig. 10 (a)). The function of RST and SF is the same as a basic pixel circuit. When a pixel is over-exposed (exposure that cannot be stored by the pixel), the excess photoelectron flows to the adjacent pixels; this is called blooming. To avoid this, the AB transistor is used. The TG transistor is used to transfer the charges from $PD$ node to the floating diffusion ($FD$) node. The DCG transistor is used to connect the storage capacitor $C_S$ when necessary. This $C_S$ is used to modify the capacitance of the $FD$ node. The current to voltage conversion gain can be controlled by the capacitance using the following equation-

$$\frac{\Delta V}{\Delta T} = \frac{I}{C_{FD} + C_S} \tag{9}$$

The ADC layer comprises a coupling capacitor ($C_C$), a comparator, a reference generator (not shown in the figure), a comparator reset switch (COMP$_{RST}$), and a state latch to control the write state of the memory. The circuit operation can be explained using the timing diagram (Fig. 10(b)). The exposure initiates in low conversion gain (LCG) mode by deactivating RST, AB, and TG transistors and activating the DCG. Concurrently, the TTS operation starts as well with applying a constant threshold voltage to the positive input terminal of the comparator. As the accumulated overflow charge on $FD$ node and the $C_S$ causes the $V_{FD}$ to cross the fixed $V_{REF}$, the comparator output state flips and the digital count value is latched in the memory. This count is the digital representation of the time needed for overflow charges to reach saturation. The next operation starts with turning off the DCG (thus disconnecting $C_S$ reduces capacitance), which enables the high gain mode, then charge transfer and auto-zeroing are performed. A single-slope ADC operation transforms the low-light signal into digital data. If this mode is not able to convert the signal because of exceeding the high gain ADC swing, the low-gain conversion intervenes by turning on the DCG transistor again; And this is called FD ADC mode. Out of 10-bit memory, digital counts from 0-511 are designated for high-gain ADC, 512-767 for FD ADC, and 768-1023 for TTS mode. Fig. 10(c) shows the light response curve. The graph clarifies which region of light intensity is converted by which mode (low lights are converted using high-gain or PD ADC mode, mid-level lights are converted through FD ADC mode, and lastly, for the brightest lights, the TTS mode is used). As this method addresses both the low and high light performance and tailors them individually in a single exposure, this would definitely widen the dynamic range. Achieving an ultra-wide dynamic range of 127 dB proves that.

## V. 3D INTEGRATION

To implement a good enough ADC inside the pixel circuit (for wide DR and high SNR), the circuitry becomes complex and accommodating the circuit per-pixel is becoming a tough challenge, and practically impossible. To overcome this situation, 3D integration (or vertical integration) is necessary. An example of 3D integration is shown in Fig. 11 (a)[71]–[74]. The whole pixel circuit is divided into multiple layers and implemented vertically. The photodetector layer can contain the photodetectors only or a small portion of simple circuit can also reside with the photo-sensitive elements. The next layers can be one or multiple ADC layers and other signal processing layers. ML algorithm layers can reside in these layers.

The layers are connected via hybrid-interconnects (Fig. 11 (b)), which connect every pixel to the next layer circuit, thus per-pixel signal-processing can be implemented. Hybrid-bonding is crucial to enable this kind of high-density connection. Hybrid-bonding refers to a permanent bond that combines a dielectric bond with embedded copper to create interconnections. This technology enables precise vertical alignment while achieving high-throughput scaling. Heterogeneous integration opens up opportunities such as depending on the specific requirements of each layer, it is possible to design each layer using a particular technology node that differs from those used in other layers and then integrated together i.e., there is no obligation for all layers to conform to the same technology node. This even allows integration between existing and emerging technologies (such as HyperFET, FerroelectricFET, Phase Change Memory, Resistive RAM, Magnetic RAM, etc.), which enables us to exploit the advantages of emerging devices for specific requirements along with the reliability of existing devices [75]. For stacking more than two chips, hybrid bonding alone may not be sufficient. Achieving multiple chip stacking requires the integration of Back-End-of-Line (BEOL) compatible transistors along with Through-Silicon Vias (TSVs). BEOL-compatible transistors allow additional active layers to be placed in the BEOL, while TSVs provide the vertical interconnects necessary for efficient communication between multiple stacked chips[76]. This technique is illustrated in Fig. 11 (c). An example of using vertical integration in pixel sensor is discussed in the next section.

### A. CROSS LAYER FEEDBACK

Mukherjee et al. utilized 3D integration to implement feedback-based bit depth variation [73]. They have used cross-layer feedback to reduce energy consumption by using lower depth for non-ROI regions. Four layers are vertically stacked (Fig. 12(a))[73]. The first layer is the photodiode layer, in which photocurrents are generated and transferred to the photocurrent to frequency (PFC) conversion layer. The generated pulses from the PFC layer are counted in the next layer (Counter layer) and the digitized values are then transferred to the ML algorithm layer. In this layer, a DNN based object detection task takes place to find the ROI of a frame. As a DNN, they have used single shot detector-based architecture coupled with Mobilenet v1 backbone. After detecting the ROI and generating predictions for the next frames, the bit depth control signals are sent as feedback. The overview of the whole mechanism is shown in Fig. 12(b).

To control the bit-depth, a circuit modification is made (inset of Fig. 12(c)), where there is a bleeder circuit that injects current to the $FD$ node. The PFC generated pulse frequency ($f_{PULSE}$) is dependent on the current of the $FD$ node according to equation (10).

$$f_{PULSE} = \frac{1}{C_{FD}} \times \frac{I_{Ph} - I_{Bld}}{\Delta V} \qquad (10)$$

Here, $I_{Ph}$, $I_{Bld}$ and $C_{FD}$ are photodiode current, bleeder current and the capacitance of the $FD$ node respectively. $\Delta V$ is the voltage difference between the reset voltage and the reference voltage of the comparator in PFC. The control signal (Ctrl) of the bleeder circuit, which comes as feedback from the ML algorithm layer, determines the current $I_{Bld}$. For different bleeder currents, the output pulse frequency will be different even if the illumination is the same. For the non-ROI pixels (ROI pixels), the $I_{Bld}$ is higher (lower), thus reducing (increasing) the resultant current, which results in a lower (higher) output pulse frequency. The graph in Fig. 12(c) shows the pixel response for different bit-depths. For the region of least interest, the maximum output is within two quantization levels, which requires only one bit for analog to digital conversion. On the other hand, for the region of maximum interest, the maximum output reaches $2^8$ quantization levels that require the full 8-bit depth. Depending on the region of other levels of interests, the bit depths (4 and 2 bits) are determined. In this way, full 8-bit depth usage can be reduced to lower energy consumption. 53-68% active energy saving is possible per frame and the ROI control adds only 9% extra energy [73].

## VI. NOISE IN DPS

The noise in a DPS can be divided into two types- random noise and pattern noise. Zhang et al. has done a thorough noise analysis of DPS [55]. Random noise varies with time and frame. It comprises shot noise and reset noise. Shot noise comes from the random carrier generation of the photodiode. The random generation of electron-hole pairs is the root of photoelectron shot noise ($v_{Ph}$) whereas thermal generation in the depletion region causes dark current shot noise ($v_d$). The $v_{Ph}$ and $v_d$ can be described using the following equations.

$$v_{Ph} = \sqrt{q \times \frac{V_{RESET} - V_{REF}}{C_{PD}}} \qquad (11)$$

$$v_d = \sqrt{q \times \frac{V_{RESET} - V_{REF}}{C_{PD}} \times \frac{I_D}{I_{Ph}}} \quad (12)$$

Here, $I_D$ and $I_{Ph}$ are dak current and photocurrent respectively. The overall shot noise is the RMS of the two shot noises. As can be seen from the equation (11) that, $v_{Ph}$ does not depend on the photocurrent (i.e., the illumination level). On the other hand, $v_d$ decreases with illumination (Fig. 13 (a)). However, the equations are derived considering the infinite integration time. In practical conditions, there will be a limitation in the integration time which implies a minimum detectable luminance ($L_{\min}$). This modifies the equations (11) and (12) as the following (which is applicable from zero illumination to $L_{\min}$)-

$$v_{Ph} = \sqrt{qL \times \frac{V_{RESET} - V_{REF}}{C_{PD} \times L_{\min}}} \quad (13)$$

$$v_d = k \times \sqrt{q \times \frac{V_{RESET} - V_{REF}}{C_{PD} \times L_{\min}}} \quad (14)$$

Here $L$ is the illumination level and $k$ is a constant. As the equations suggest, for actual conditions, $v_{Ph}$ rises with increasing luminance up to $L_{min}$ and on the other hand, $v_d$ remains constant. The dashed lines in Fig. 13 (a) show the actual conditions.

There are two types of pattern noises in DPS. The first one is pixel fixed pattern noise (PFPN), and the other is photo-response non-uniformity (PRNU). The latter one is ignorable because it has nearly zero impact on the image quality. Noise introduced by comparator offset is the main source of pattern noise in a DPS.

Temporal noise with varying capacitance and reference voltage is shown in Fig. 13 (b) [55]. Noise decreases with increasing capacitance, but comparator offset error also increases, thus optimization is required according to the application. Increasing the reference voltage, reduces noise but that will limit the swing ($V_{RESET}$ to $V_{REF}$) of $V_{PD}$. So, there is a trade-off here as well and the choice should be made according to the requirements of the application.

## VII. COMPARISON OF DIGITAL PIXEL SENSORS

Due to scaling, the supply voltage is decreasing. The voltage-domain based conversions (SAR ADC, MCBS, etc.) depend on the supply voltage whereas time-domain conversion (PWM, PFM, etc.) does not. Thus time-domain conversions are suitable for scaling. Table III shows a comparison of different parameters among different DPSs and some other types of pixel sensors [33], [77]–[80]. When comparing the pixel size, we must keep in mind that it will not only depend on circuit complexity but also on the CMOS technology used. We can see improvements from [65] to [81] in several parameters, because in [81], the delta-sigma modulator is shared among four pixels, however, this deteriorates the DR. In [61], the power consumption is low, and the DR is high.

The reason behind low power is the decreased output bit rate by using a recursive method in the decimator (off-pixel). But in this case, high SNR cannot be achieved due to a limited OSR. A free-running continuous oscillator which is sampled at fixed intervals, offers high DR. In [82], a very low pixel size is achieved because they used the photodiode itself as an integrator (which is an important component of delta-sigma modulator) without using any external one. The power consumption of this sensor seems very low, but this power is reported without the decimator power consumption. In [83], Mahmoodi et al. designed the sensor circuit keeping the decimator inside the pixel, which results in a low fill factor and high-power consumption but a very high DR. [58] shows the Nyquist rate MCBS. Unlike the delta-sigma modulator, this does not require oversampling. [57] presented a SAR type ADC based DPS, it requires a huge amount of energy per pixel, which is used for special purpose laboratory applications such as particle capturing. The next five DPSs ([69], [84], [85], [35], [86]) in the table are PWM based. Among them, in [35], instead of the traditional comparator, they used a simple mixed-mode circuit, which operates at a low supply voltage and thus consumes a very low power. This circuit does not involve any bias reference circuit. [39] and [68] are PFM based pixel circuits, usually the DR range is high for this kind of conversion. In [68], they used an in-pixel variable reference generation circuit. This scheme pushes the low light detection limit deeper by increasing the reference voltage when low illumination is detected and thus achieving an ultra-wide dynamic range of 150 dB. [63] is an example of 3D integration. A very small pixel pitch became possible for vertical stacking, and this allows a high-density pixel array. The table also presents performance metrics for analog and passive pixel sensors, providing a clear comparison between Digital Pixel Sensors and other imaging circuits. Fig. 14 shows a timeline of some of the most major milestones in the journey of digital pixel sensors.

## VIII. APPLICATIONS OF DPS

Digital Pixel Sensor (DPS) technology is particularly beneficial for applications that require high speed and low noise performance (as the digital signal traversing the array is noise-immune), and can accommodate a larger, bulkier size [67], [87]. This technology is being applied in numerous fields, each benefiting from the unique advantages that DPS brings over traditional analog pixel sensors. Here, we explore several key applications of DPS technology and discuss new and emerging areas where DPS could significantly impact.

DPS benefits scientific imaging applications such as microscopy [88]. Each pixel in a DPS is capable of detecting light with high precision, which is crucial for observing fine details in microscopic samples. The low noise characteristics of DPS ensure that the images are clear and free from artifacts, enabling accurate analysis and measurements. DPSs are used in space telescopes and other astronomical instruments to capture high-resolution images of celestial bodies, benefiting from reduced noise and higher sensitivity [89]. Surveillance cameras equipped with DPS benefit significantly in terms of image clarity across diverse lighting conditions, thereby enhancing security monitoring [86]. The localized conversion of DPS allows for real-time adaptation to varying light levels, ensuring that images remain clear and detailed whether in bright daylight or low-light scenarios. The ability to maintain high image quality in different lighting conditions enables security systems to accurately capture and interpret crucial details, such as facial features or license plates, thereby increasing the effectiveness and reliability of surveillance operations. Autonomous vehicles can utilize cameras with DPS to achieve real-time, low-noise imaging, which significantly enhances safety and navigation [90]. DPS enables the vehicle to detect and respond to obstacles, traffic signals, pedestrians, and other critical elements in its surroundings with greater precision. The rapid and accurate interpretation of visual data provided by DPS is essential for making split-second decisions, thereby improving the overall safety and efficiency of autonomous navigation.

## IX. CURRENT & FUTURE TRENDS

Based on the discussion of the previous sections, some trends in the field of digital pixel sensors are clearly visible, and the trajectory also guides the future directions. The technology will embrace the efficiency of tailored (ADC) resolutions. Recognizing that not every detail in a scene holds equal significance, this approach optimizes resources by allocating higher ADC resolutions only to pixels where intricate information is vital while employing lower resolutions elsewhere [73]. This adaptive ADC strategy promises enhanced computational efficiency and reduced power consumption, paving the way for more sophisticated and resource-conscious imaging systems. Event-driven imaging is another aspect of future trends in the DPS research area, where ADC conversion is triggered only when significant changes occur in the scene rather than continuously sampling at a fixed rate. This approach reduces power consumption and data bandwidth while enabling high-speed capture of dynamic scenes. Sensor fusion can emerge as a pivotal approach, which involves integrating multiple types of imagers, such as those capturing visible light and infrared wavelengths, to enhance the robustness of decision-making processes. Vertical integration is going to be an inevitable replacement for traditional planar integration. Using vertical integration, multiple ML layers can be added. This enables image sensors to not only capture data but also to analyze and process it on-device in real-time. By harnessing ML algorithms within the sensor itself, efficiency is maximized as data doesn't need to be transferred elsewhere for processing [75]. This streamlined approach enhances response times and reduces the computational burden on external systems. Different emerging devices that are currently being used in neuromorphic[91], memory circuits[92], etc., can be tailored for image sensor circuits to achieve low-power and high-speed imaging[93], [94]. In summary, innovative approaches in DPS technology promise heightened efficiency, reduced power consumption, and real-time processing capabilities, ushering in a new era of sophisticated and resource-conscious imaging systems.

## X. CONCLUSION

For fighting the scaling issues and achieving high speed, the digital pixel sensor has no alternative. Processing within individual pixel becomes a necessity and digital pixel sensor paves the way for it. However, the ADC inside the pixel circuit makes the pixel very complex and area hungry. So, there is an opportunity for research and development to make more compact conversion circuits while maintaining good performance such as high DR, pixel density, fill factor, SNR, and maintain low power consumption and noise. Currently, we are standing at the intersection point of technological progress and creative potential. The voyage of digital pixel sensors is far from over. In this era of artificial intelligence, the DPS can aid in applying pixel-level machine learning-based algorithms. This review article explored the multifaceted realm of digital pixel sensors, and shed light on their operation mechanisms, pros, and cons, etc. A comparative study among different types of DPSs has also been presented. The overall compilation of this review can provide a valuable foundation for paving the way towards new research avenues in the widespread field of digital pixel sensors.

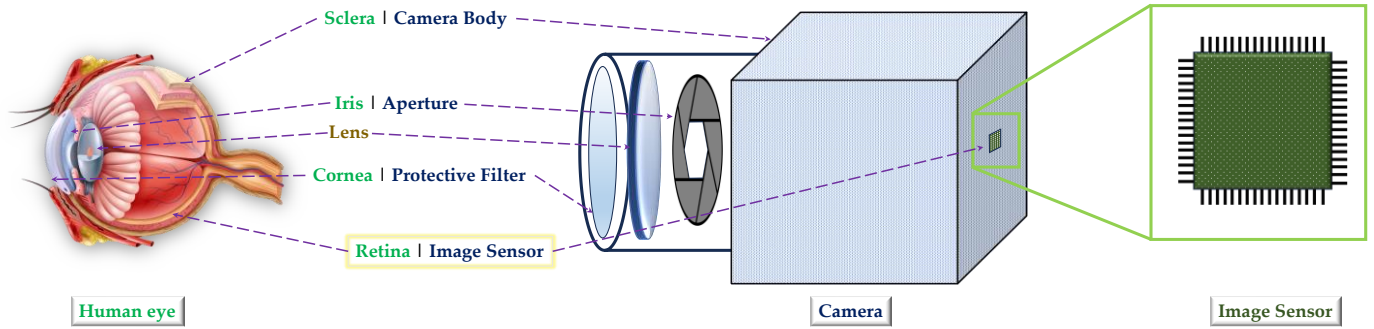

FIGURE 1. Similarities between a human eye and a basic camera. Iris/Aperture controls the amount of light that can enter into the eye or camera. Single or multiple lenses are used to control focus. The image sensor chip (which comprises photodetectors, readout circuits, ADCs, etc.) converts light energy to electrical signals imitating the function of the retina. The image sensor is the most sophisticated part of the camera. [14]

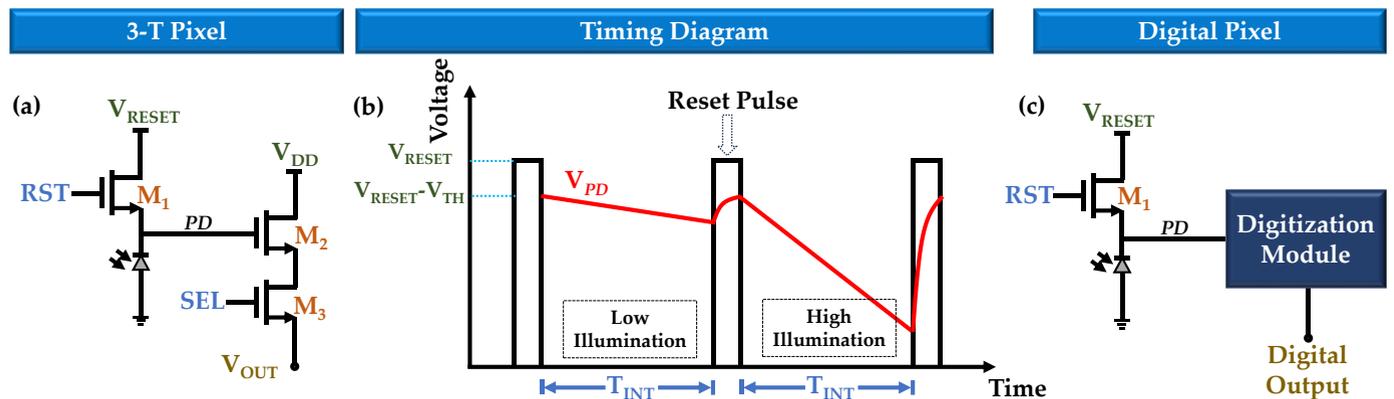

FIGURE 2. (a) Basic 3-T pixel circuit diagram. $M_1$, $M_2$, and $M_3$ are reset, source follower, and row selector transistors, respectively. (b) The timing diagram of a basic 3-T pixel. Operation starts with enabling the reset transistor; the PD node voltage becomes $V_{RESET}$-$V_{TH}$, where $V_{TH}$ is the reset transistor threshold voltage. By turning off the reset transistor, the integration period begins. Light falling on the photodiode causes a voltage drop at the node PD. This voltage drop is faster in case of higher illumination. At the last part of the cycle, the voltage drop in a predefined integration time ($T_{INT}$) at the node PD is read out by enabling $M_3$. The read-out voltage is converted to a digital signal outside the pixel. (c) Digital pixel, where digitization takes place inside the pixel circuit.

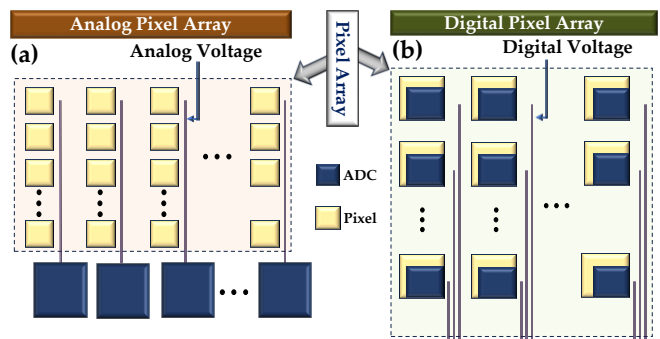

FIGURE 3. (a) Analog pixel sensor array. Analog to digital conversion takes place at the end of each column. Each digitization module processes one row at a time; hence it is time-consuming. (b) Digital pixel sensor array. Digitization takes place inside the pixel. Though pixels are bulkier, this process can achieve higher speed.

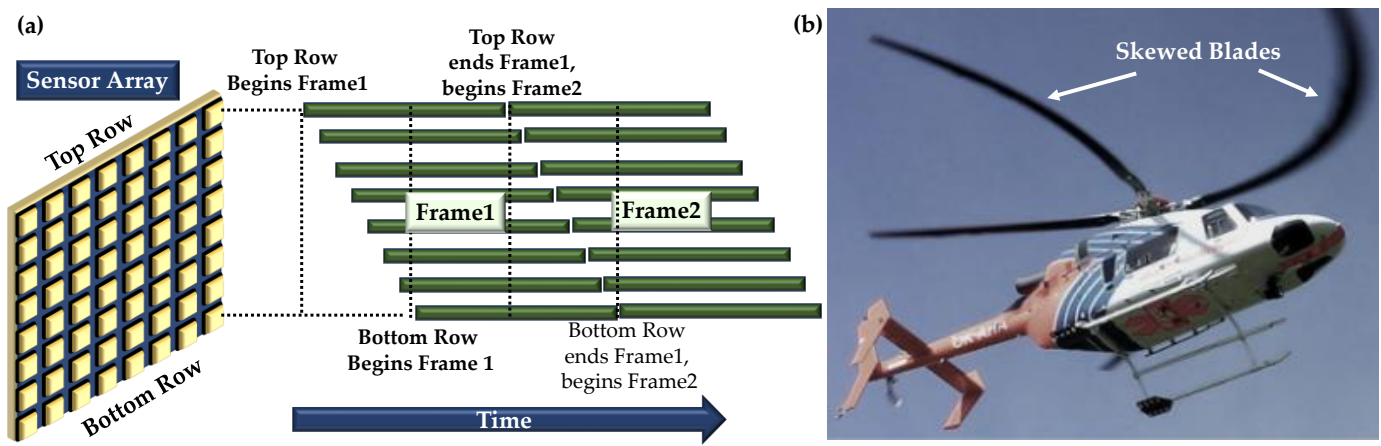

FIGURE 4. (a) Rolling shutter in an analog pixel sensor. At first, the top row is read out, then sequentially it moves to the bottom row. There is a time difference between each row readout. So different parts of a frame are captured at different times. If there is a relative motion between the camera and the scene, the image will be distorted. (b) A sample image with the rolling shutter effect. Although the blades are straight in real life, as they are in motion, they look skewed due to the rolling shutter effect.

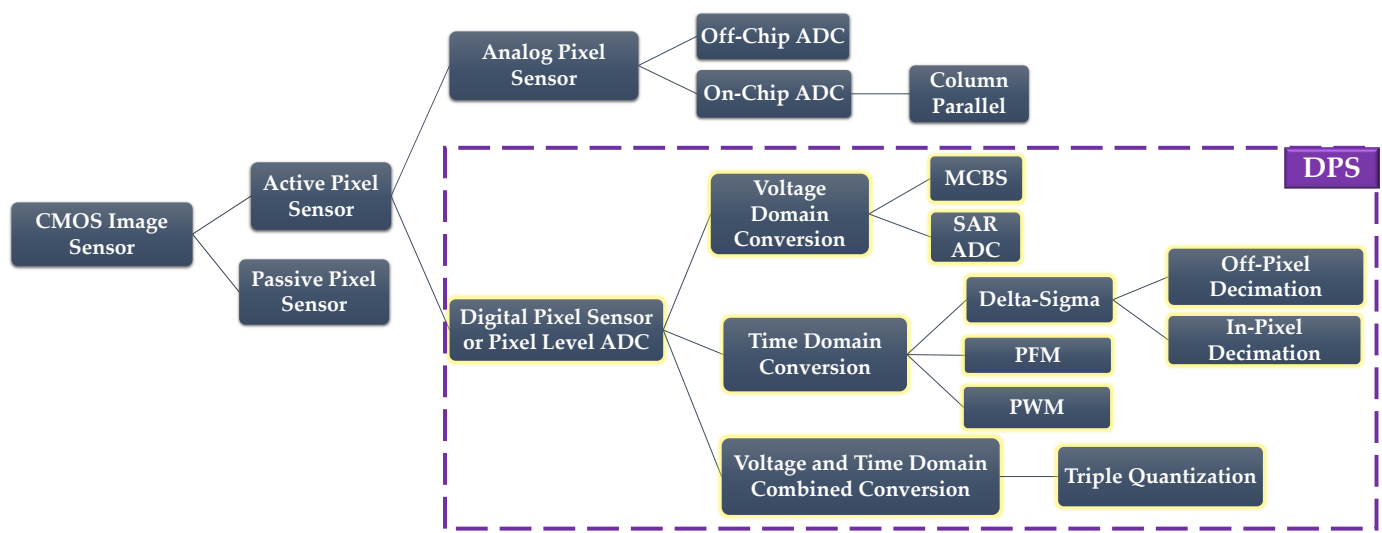

FIGURE 5. Classification of CMOS image sensors based on where and how the ADC is taking place. The focus of this paper will be on the digital pixel sensors (DPS), where ADC takes place inside the pixel circuit. DPSs are divided into different categories based on which domain is being used to digitize the light signal.

TABLE I
**Gray Code Quantization**

| Analog voltage range (V) | Gray code |
|---|---|
| 0-0.125 | 000 |
| 0.125-0.25 | 001 |
| 0.25-0.375 | 011 |
| 0.375-0.5 | 010 |
| 0.5-0.625 | 110 |
| 0.625-0.75 | 111 |
| 0.75-0.875 | 101 |
| 0.875-1 | 100 |

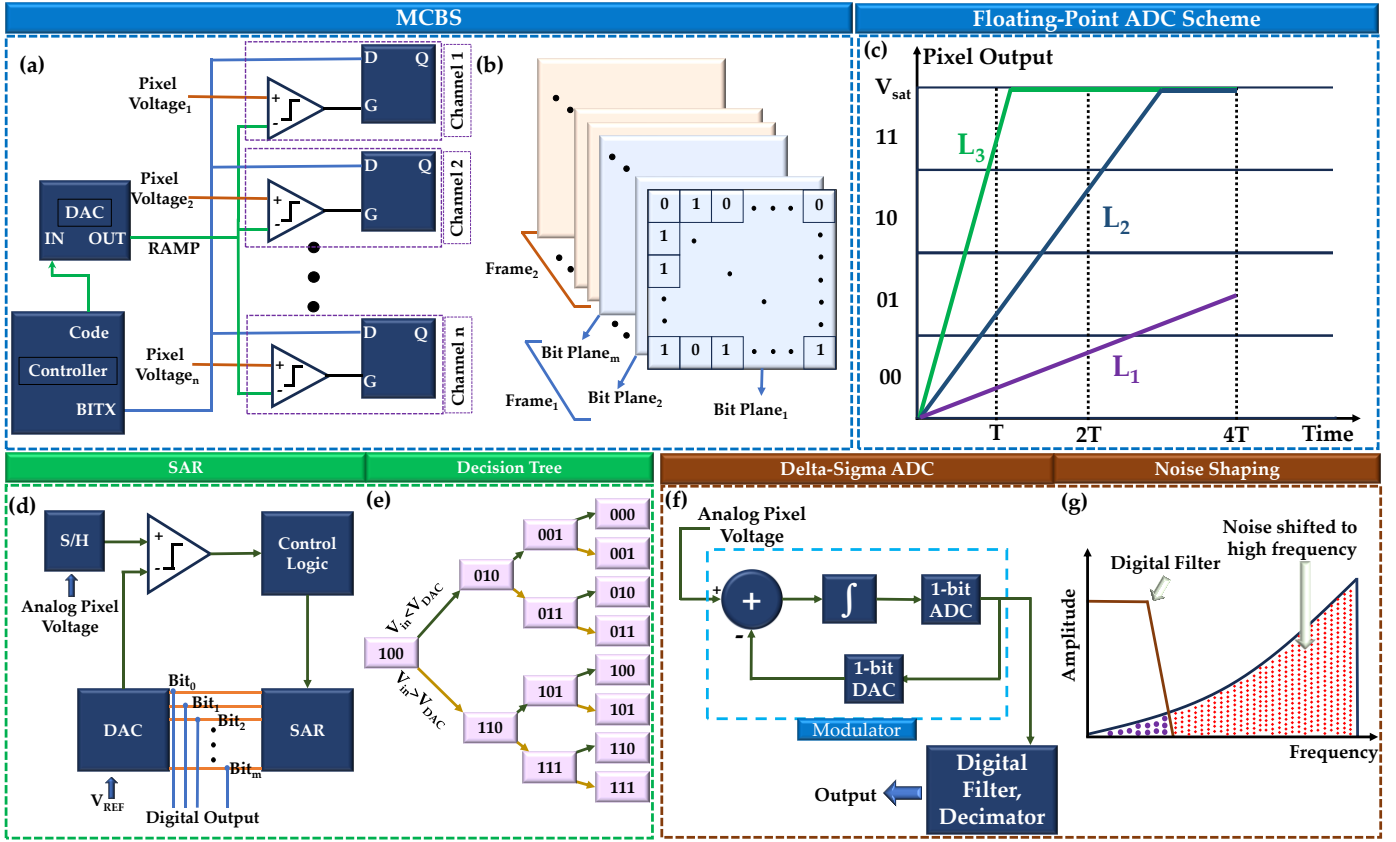

FIGURE 6. (a) MCBS digitization block diagram. Each channel consists of a comparator and a latch; the comparator takes pixel voltage as an input. The DAC and controller module are shared among the channels. By taking code word input from the controller block, DAC generates an increasing staircase-shaped reference voltage (RAMP) for the comparator. BITX represents the individual bit, which toggles whenever the RAMP changes and latches if the comparator flips. (b) All the channels in a 2D array generate a bit plane, and m number of bit planes form a frame. The m number of bits of a particular coordinate of the array represents the digital representation of the light intensity of that position. (c) Pixel output and digitized codes vs. time. $L_1$, $L_2$, and $L_3$ represent different levels of illumination in 3 different pixels. At T, 2T and 4T time, 3 sampling operations are performed. (d) SAR type ADC. Analog pixel voltage is sampled and held in the S/H block to feed the positive terminal of the comparator. The comparator compares the pixel voltage with the DAC-generated reference. m number of comparisons are done successively for m bit depth. (e) The decision tree of a 3-bit SAR type ADC. 3-bit binary numbers in the figure represent the output of the registers. (f) Delta-sigma ADC based DPS. The difference or error between the pixel voltage and the output of the DAC is passed to the integrator and then to the 1-bit ADC. The oversampled signal is averaged and decimated in the next block. As the quantization noise is shifted to high frequency, a digital low-pass filter is used to attenuate the noise. (g) Graphical representation of noise shaping. The quantization noise shifted to high frequency is filtered out using a digital filter.

**TABLE II**
**Digitization with floating-point number**

| Illumination Level | b1 | b2 | b3 | b4 | Exponent | Mantissa |
|---|---|---|---|---|---|---|
| $L_1$ | 0 | 0 | 0 | 1 | 0 | 01 |
| $L_2$ | 0 | 1 | 0 | 1 | 1 | 10 |
| $L_3$ | 1 | 1 | 1 | 1 | 2 | 11 |

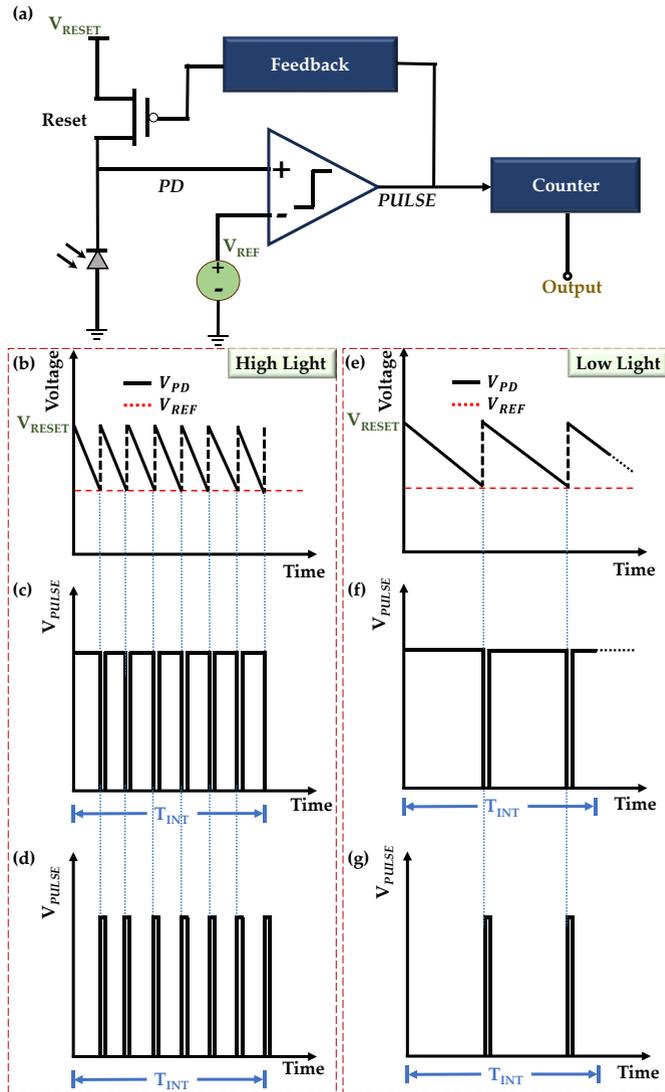

FIGURE 7. (a) PFM based pixel circuit diagram. The comparator compares the voltage between the PD node and the predefined reference voltage $V_{REF}$. When $V_{PD}$ crosses $V_{REF}$, the comparator output flips, generates a pulse, and turns on the reset transistor to reset the PD node voltage to $V_{RESET}$ through the feedback module. The counter counts the number of pulses in a defined time $T_{INT}$. (b) PD node voltage and the reference voltage vs. time for high illumination. (c) Comparator output vs. time for high illumination when the PD node is connected to the positive terminal and $V_{REF}$ is connected to the negative terminal of the comparator. (d) Comparator output vs. time for high illumination when the PD node is connected to the negative terminal and $V_{REF}$ is connected to the positive terminal of the comparator. (e), (f), and (g) illustrate the same graphs as (b), (c), and (d) respectively for lower illumination. In case of lower illumination, the number of pulses is lower in a predefined integration time $T_{INT}$. So, the number of pulses in a predefined time is proportional to the light intensity.

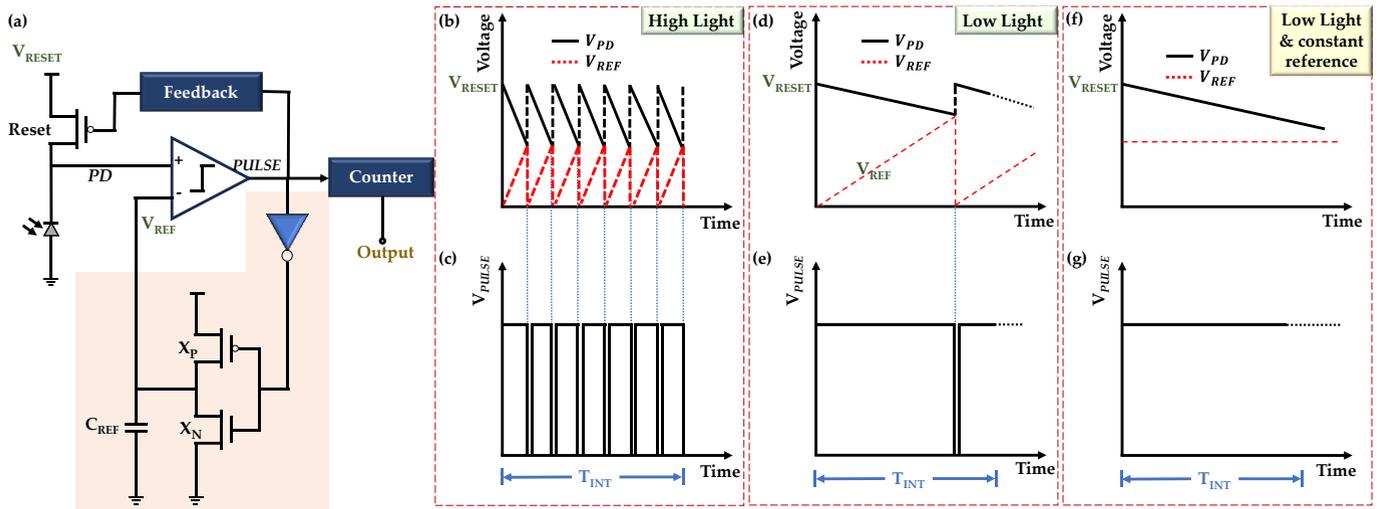

FIGURE 8. (a) In-pixel variable reference PFM circuit diagram. In addition to a basic PFM pixel circuit, it has a variable reference generation mechanism (highlighted) which consists of two inverters and a capacitor $C_{REF}$. (b) The voltage at the PD node and the variable reference vs. time graph at a high light intensity. (c) Generated pulse at the output of the comparator in the predefined integration time $T_{INT}$ at a high light intensity. (d) The voltage at the PD node and the variable reference vs. time graph at a lower light intensity. (e) Generated pulse at the output of the comparator in the predefined integration time $T_{INT}$ at a lower light intensity. (f) The voltage at the PD node vs. time graph at the same low light intensity if a fixed reference voltage was used. (g) No pulses are detected within the same time $T_{INT}$. This shows the inability of a constant reference PFM to detect a certain amount of low light.

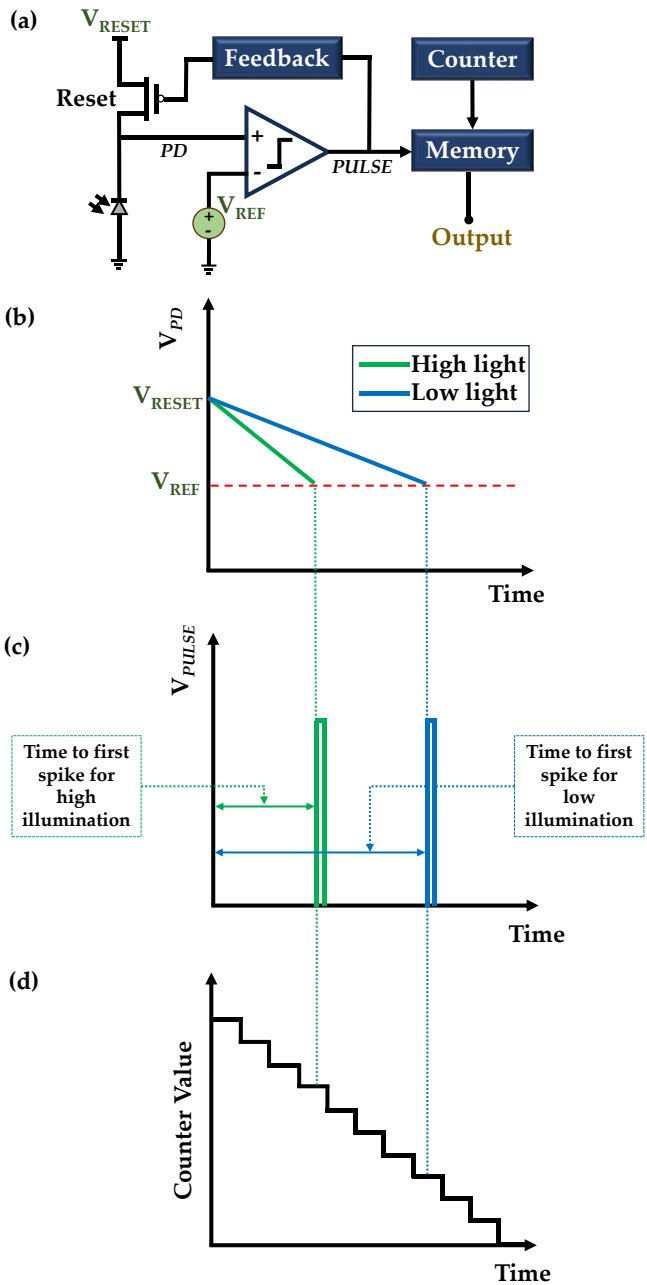

FIGURE 9. (a) PWM based pixel circuit diagram. The comparator compares the voltage between the PD node and the predefined reference voltage $V_{REF}$. When $V_{PD}$ crosses $V_{REF}$, the comparator output flips, generates a pulse, and turns on the reset transistor to reset the PD node voltage to $V_{RESET}$ through the feedback module. A down counter counts the time of $V_{PD}$ to reach $V_{REF}$ from the initial $V_{RESET}$ voltage. This count value is latched in the memory. (b) PD node voltage and the reference voltage vs. time for high and low illumination. (c) Comparator output vs. time for high and low illumination. It takes less time to get the spike for higher light intensity. (d) The count value of the down counter indicates the level of light intensity, which is latched in the memory.

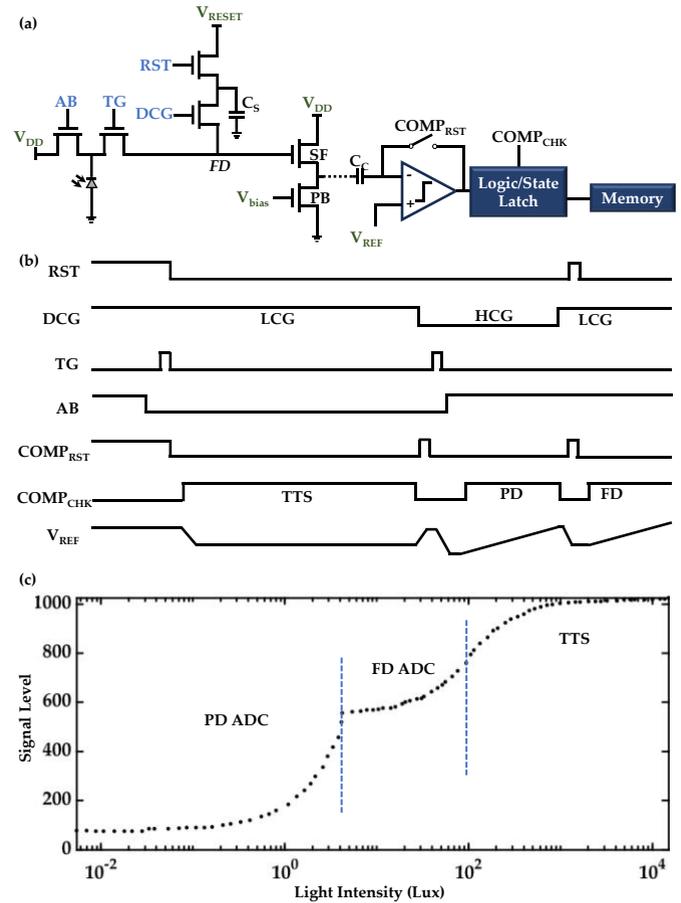

FIGURE 10. (a) Triple quantization (3Q) pixel circuit diagram. Besides the common three transistors of a pixel circuit, the other transistors are antiblooming (AB), transfer (TG), dual conversion gain (DCG), and pixel bias (PB) transistors. A storage capacitor ($C_S$) is connected to the FD node through the DCG transistor to control the capacitance of this node and thus controls the conversion gain. This portion of the circuit lies on the first layer of 3D integration and the next layer (ADC) is connected via hybrid bonding. The ADC comprises a comparator, a coupling capacitor ($C_C$), and $COMP_{RST}$ switch for CDS operation, and a latch to write the digitized value in the memory. (b) Timing diagram of 3Q operation. The exposure operation begins with the LCG mode by turning off the RST, AB, and TG and turning on the DCG; parallelly, the time to saturation (TTS) operation begins and the constant version of the reference voltage ($V_{REF}$) is fed into the comparator. When the voltage at the FD node crosses the constant $V_{REF}$, the comparator flips, and the digital count value is latched in the memory. Then the HCG operation starts with disabling the DCG. Low light analog to digital conversion is done using a single slope ADC operation (PD ADC operation). Lastly, again the LCG mode is activated to digitize the signal that exceeds the HCG swing (FD ADC operation). (c) Photo response curve. Low lights are digitized using PD ADC mode. Moderate intensity lights are converted through the FD ADC mode. The brightest lights are digitized using the time to saturation mode. [52]

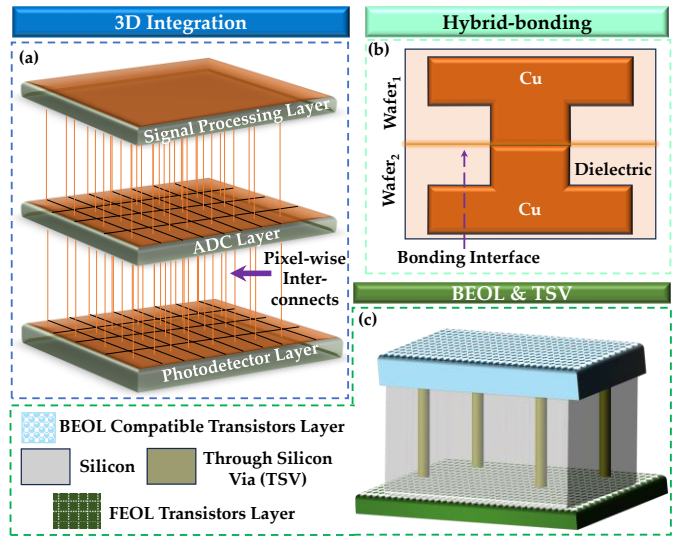

FIGURE 11. (a) 3D integration of pixel circuits where multiple layers are stacked vertically. Light hits the photo detector layer. The pixel voltage from each pixel is transferred to the ADC layer via the pixel-wise hybrid interconnects. For subsequent processing, further signal processing layers are stacked where per-pixel processing is possible. (b) A simplified view of the hybrid bonding between two wafers, where two copper pads are connected through direct bonding. This allows a high-density pixel array. (c) Back-end-of-line (BEOL) compatible transistors are placed vertically above front-end-of-line (FEOL) transistors using through-silicon vias (TSVs). This fabrication technique is crucial to achieve multiple chip stacking.

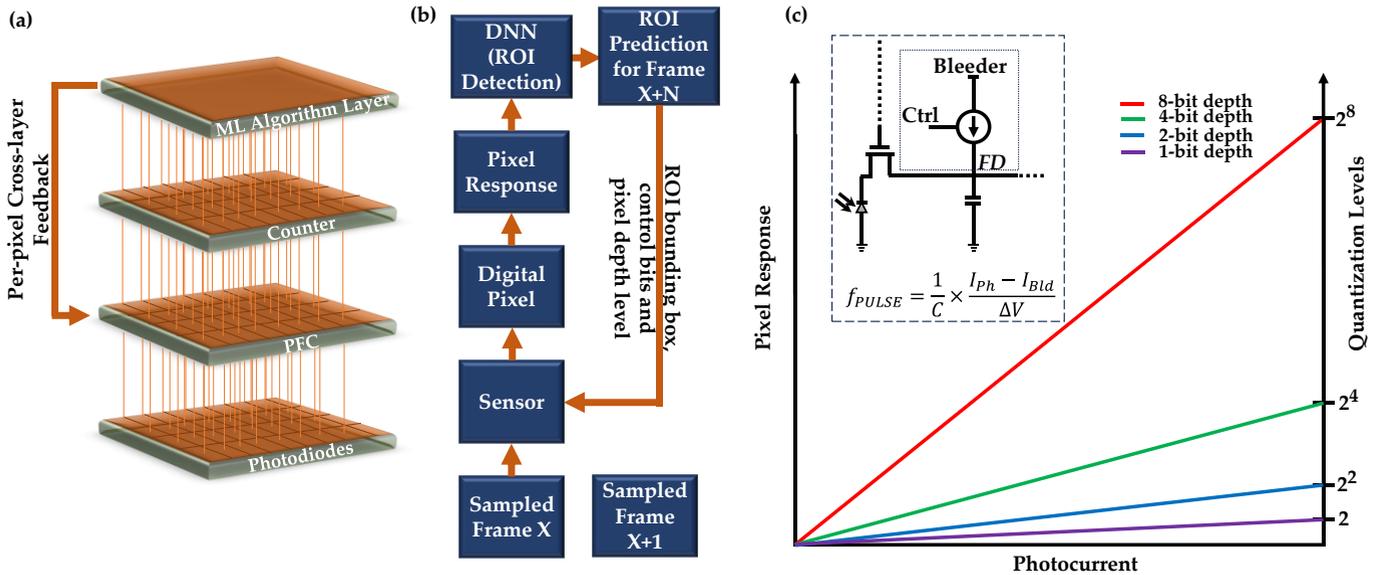

FIGURE 12. (a) 3D integrated DPS with per-pixel cross-layer feedback. For ADC, the photo-current to frequency (PFC) conversion technique is used. The counter circuits are big in size, hence separated from the PFC to another layer. The digitized value is then sent to the ML algorithm layer. (b) Overview of the mechanism. The first frame (Frame X) is sensed and transferred to the DNN layer through different conversion layers with the maximum possible quality. The DNN layer detects the ROI and generates predictions for the next layers. The ROI predictor sends control signals as feedback to the ADC to control the bit depth for the next sampled frames. (c) The inset shows the additional bleeder circuit to a usual pixel circuit. This bleeder uses the feedback from the ROI predictor and injects current to the FD node to manipulate the resultant current which is $I_{Ph} - I_{Bld}$, where $I_{Ph}$ and $I_{Bld}$ are the photodiode and the bleeder currents, respectively. The graph shows four levels of pixel-depth control. For the regions of least interest, $I_{Bld}$ is set to the highest value, which results in the lowest output pulse frequency which can swing only from 0 to 2 quantization levels, so in this case, 1 bit depth is enough. For the regions of highest interest, $I_{Bld}$ is set to the lowest value, which results in the highest output pulse frequency which can swing from 0 to $2^8$ quantization levels, so in this case, the highest 8-bit depth is used.

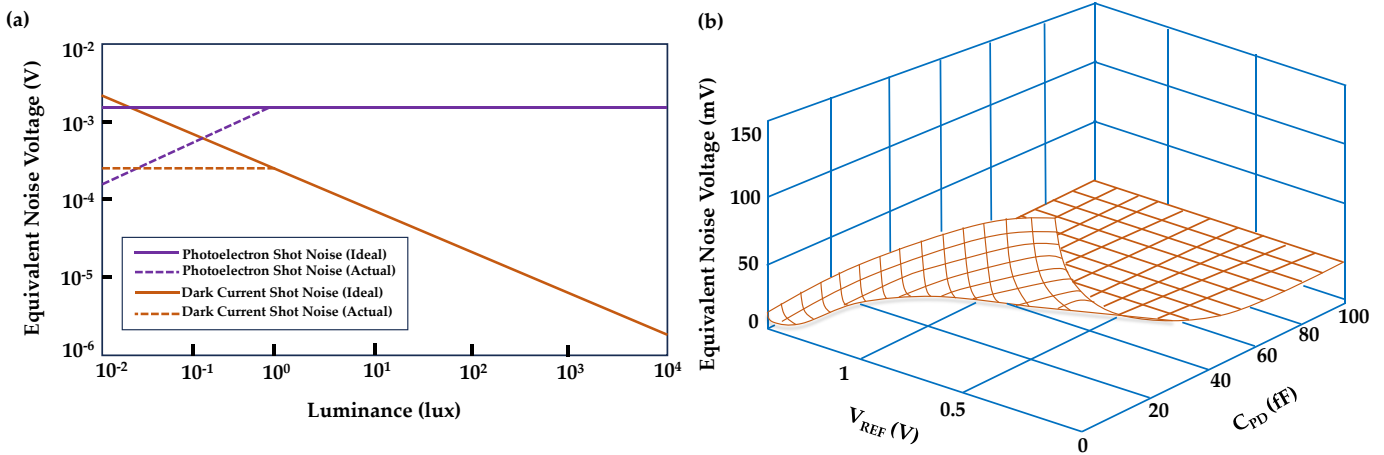

FIGURE 13. (a) Shot noise vs. illumination in a PWM DPS. According to equation (11) ideal photodiode shot noise is constant with respect to luminance and according to equation (12) ideal dark current shot noise decreases with illumination level. However, there is a limit of the highest integration time in a DPS, which corresponds to a minimum illumination $L_{min}$ that can be detected. Up to this $L_{min}$ point, the noise behavior follows equations (13) and (14), the photodiode shot noise becomes proportional to luminance and the dark current shot noise remains constant. (b) Temporal noise vs. photodiode capacitance and the reference voltage for a PWM DPS. High capacitance reduces the noise but results in an increased comparator offset error. The higher reference voltage improves noise performance but limits the voltage swing between the reset voltage and the reference voltage. So, an optimization should be made according to the requirements set by the application. [55]

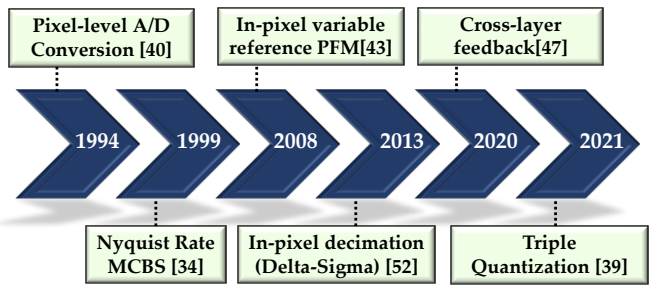

FIGURE 14. A timeline of digital pixel sensor evolution, showing milestones.

TABLE III
Comparison of different DPSs along with some other types of pixel sensors.

| Specs. Ref. | Type of ADC | CMOS Technology ($\mu m$) | Pixel Size ($\mu m^2$) | Resolution or Array Size (pix×pix) | Fill Factor | Supply Voltage (V) | Frame Rate (frames per second) | DR (dB) | Power Consumption (nW/pixel) |
|---|---|---|---|---|---|---|---|---|---|
| [65] | Delta-Sigma | 1.2 | 60×60 | 64×64 | 3% | 5 | - | 93 | <244 |
| [81] | Delta-Sigma | 0.8 | 20.8×19.8 | 128×128 | 30% | 3.3 | - | 83 | <61 |
| [61] | Delta-Sigma | 0.5 | 30×30 | 48×48 & 64×64 | - | 3.3 | - | 104 | 40 |
| [82] | Delta-Sigma | 0.35 | 10×10 | 128×128 | 31% | 3.3 | - | 99.6 | 0.88 (w/o decimator) |
| [83] | Delta-Sigma (in-pixel decimation) | 0.18 | 38×38 | 48×64 | - | 1.8 | - | >110 | 1399 (analog power*) & 3645 (digital power*) |
| [58] | MCBS | 0.35 | 10.5×10.5 | 320×256 | 28% | 3.3 | - | - | 240 |
| [57] | SAR | 0.18 | 35×35 | 16×16 | - | 1.8 | 100 ns/row | - | 140 $\mu W$/pixel |
| [69] | PWM | 0.18 | 12×14 | 4×4 | 13% | 1.2 | - | - | 180 |
| [84] | PWM | 0.35 | 45×45 | 64×64 | 12% | 3.3 | - | 85 | - |
| [85] | PWM | 0.35 | 45×45 | 64×64 | - | 3.3 | - | >100 | - |
| [35] | PWM | 0.18 | 21×21 | 16×16 | 39% | 0.8-1.8 | 5000 | 140 | 2.85 |
| [86] | PWM | 0.18 | 9.4×9.4 | 352×288 | 15 | 2.1 | 10000 | - | ~ 493 |
| [39] | PFM | 0.18 | 23×23 | 28×28 | 25% | 0.8-1.2 | - | 130 | 250 |
| [68] | PFM (in-pixel variable reference) | 0.18 | - | - | - | 1.8 | - | 150 | - |
| [63] | 3D integration and triple quantization | 45 nm | 4.6×4.6 | 512×512 | - | - | 480 | 127 | 20.2 |
| [†]Analog Pixel Sensor [77] | | 2 | 40×40 | 128×128 | 26% | 5 | 30 | 71 | - |
| [†]Analog Pixel Sensor [78] | | 0.18 | 10×10 | 7×2 | - | 1.8-3.3 | - | - | 0.39 mW (only gain amplifier) |
| [†]Analog Pixel Sensor [33] | | 0.18 | 3.63×3.63 | 1920×1440 | - | 1.8-3.3 | 180 | 68 | 580 mW (whole chip) |
| [†]Analog Pixel Sensor [79] | | 0.11 | 2.9×2.9 | 1024×240 | - | 1.8-2.8 | 570 | 73.6 | 57.2 mW (whole chip) |
| Passive Pixel Sensor [80] | | 0.6 | 20×20 | 256×256 | 53% | 3.3 | 30 | - | 10.5 mW (whole chip) |

*Analog power refers to sensor and modulator power, whereas digital power means decimator and readout power.

[†]Analog signal from each pixel traversing through the array.

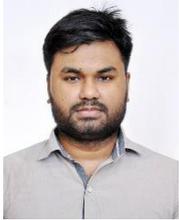

**MD RAHATUL ISLAM UDOY** received the B.S. degree in Electrical and Electronic Engineering from Bangladesh University of Engineering and Technology in 2019. He is currently pursuing a Ph.D. degree in Electrical Engineering at the University of Tennessee Knoxville, TN, USA. Prior to beginning his graduate studies, he worked in the Daffodil International University (Bangladesh) as a full-time lecturer.

From July 2022, he is a Research Assistant with the Nanoelectronic Devices and Integrated Circuits (NorDIC) Lab. His research interests include in-sensor computing, beyond CMOS technologies, atomistic simulation, device modeling, circuit design, etc.

Mr. Udoy's awards and honors include the Tennessee's Top 100 Graduate Fellowship, Bangladesh-Sweden Trust Fund Grant, and DAC Young Fellowship.

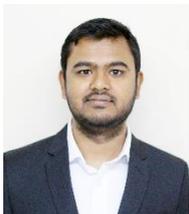

**SHAMIUL ALAM** received the B.S. degree in Electrical and Electronic Engineering from Bangladesh University of Engineering and Technology in 2017. He is currently pursuing a Ph.D. degree in Electrical Engineering at the University of Tennessee Knoxville, TN, USA.

From January 2020, he is a Research Assistant with the Nanoelectronic Devices and Integrated Circuits (NorDIC) Lab. His research interests include device modeling and circuit design for logic, memory, and in-memory computing applications.

Mr. Alam's awards and honors include the Tennessee's Top 100 Graduate Fellowship, DAC Young Fellowship, Graduate Advancement Training and Education (GATE) Fellowship, and Gonzalez Family Award for Outstanding Graduate Research Assistant.

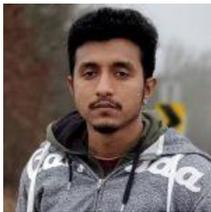

**Md Mazharul Islam** received his B.S. in Electrical and Electronic Engineering from Bangladesh University of Engineering and Technology in 2017. He is currently pursuing a Ph.D. degree in Electrical Engineering at the University of Tennessee Knoxville, TN, USA.

From August 2020, he is a Research Assistant with the Nanoelectronic Devices and Integrated Circuits (NorDIC) Lab. His research interests include beyond CMOS technologies, neuromorphic computing, etc.

Mr. Islam's awards and honors include the Tennessee's Top 100 Graduate Fellowship and DAC Young Fellowship.

.

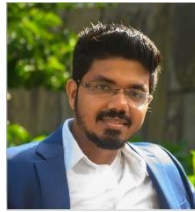

**Akhilesh Jaiswal** is an Assistant Professor of Electrical and Computer Engineering at the University of Wisconsin-Madison. He obtained his Ph.D. in from Purdue University in 2019. Prior to that, he received his M.S. from University of Minnesota-Minneapolis in 2014 and B.S. from Shri Guru Gobind Singhji Institute of Engineering and Technology, Nanded, India in 2011.

His current research interests include intelligence at extreme-edge using processing-in-pixel and processing-in-sensor technology, hardware-software co-design for efficient distributed computing, bio-inspired paradigms for sensing and compute including neuromorphic systems, electro-optic general-purpose computing among others.

Dr. Jaiswal's awards include ISI Exploratory Research Award in 2020, Keston Exploratory Research Award in 2021, IEEE Brain Community Best Paper Award in 2022, and Best Paper Nomination in VLSI-SoC 2022.

**AHMEDULLAH AZIZ** is an Assistant Professor in the Dept. of EECS at the University of Tennessee, Knoxville. He received his Ph.D. in Electrical and Computer Engineering from Purdue University in Fall 2019. He earned an M.S. degree in Electrical Engineering from Pennsylvania State University (University Park) in 2016 and a B.S. degree in Electrical & Electronic Engineering from Bangladesh University of Engineering & Technology (BUET) in 2013.

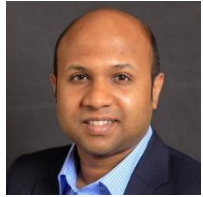

Prior to beginning his graduate studies, he worked in the 'Tizen Lab' of Samsung R&D Institute as a full-time Engineer. He also worked as a Co-Op Engineer (Intern) in the Technology Research division of Global Foundries (Fab 8, NY, USA). His research interests include mixed-signal VLSI circuits, non-volatile memory, and beyond CMOS device design.

Dr. Aziz received several awards and accolades for his research, including the 'Outstanding Dissertation Award' from European Design and Automation Association (2020), 'Outstanding Graduate Student Research Award' from College of Engineering, Purdue University (2019) and 'Icon' award from Samsung (2013). He was a co-recipient of two best publication awards from SRC-DARPA STARnet Center (2015, 2016) and best project award from CNSER (2013).